\pdfoutput=1

\documentclass[11pt]{article}

\usepackage[final]{ACL2023}

\usepackage{times}
\usepackage{latexsym}

\usepackage[T1]{fontenc}

\usepackage[utf8]{inputenc}

\usepackage{microtype}

\usepackage{inconsolata}
\usepackage{amsfonts}
\usepackage{amsmath}
\usepackage{amssymb}       
\usepackage{bold-extra}    
\usepackage{bm}            
\usepackage{graphicx}      
\usepackage{multirow}      
\usepackage{booktabs}      
\usepackage{enumitem}      
\usepackage{subcaption}    
\usepackage{pgfplots}
\usepackage{pgf-pie}  
\usepackage{tikz}
\pgfplotsset{compat=newest}
\usepackage[hang,flushmargin]{footmisc}  

\title{Do We Still Need Audio? Rethinking Speaker Diarization with a Text-Based Approach Using Multiple Prediction Models}

\author{Peilin Wu \\
  Computer Science \\
  Emory University \\
  Atlanta, GA 30322 USA \\
  \texttt{\small peilin.wu@emory.edu} \\\And
  Jinho D. Choi \\
  Computer Science \\
  Emory University \\
  Atlanta, GA 30322 USA \\
  \texttt{\small jinho.choi@emory.edu} \\}

\begin{document}
\maketitle
\begin{abstract}
We present a novel approach to Speaker Diarization (SD) by leveraging text-based methods focused on Sentence-level Speaker Change Detection within dialogues. Unlike audio-based SD systems, which are often challenged by audio quality and speaker similarity, our approach utilizes the dialogue transcript alone. Two models are developed: the Single Prediction Model (SPM) and the Multiple Prediction Model (MPM), both of which demonstrate significant improvements in identifying speaker changes, particularly in short conversations. Our findings, based on a curated dataset encompassing diverse conversational scenarios, reveal that the text-based SD approach, especially the MPM, performs competitively against state-of-the-art audio-based SD systems, with superior performance in short conversational contexts. This paper not only showcases the potential of leveraging linguistic features for SD but also highlights the importance of integrating semantic understanding into SD systems, opening avenues for future research in multimodal and semantic feature-based diarization.
\end{abstract}

\section{Introduction}
Speaker Diarization (SD), essential for enhancing Speech-To-Text systems, identifies "who speaks what" in audio conversations jointly with Automatic Speech Recognition (ASR) systems. This technology is crucial for various recording applications and preparing AI training datasets, where distinguishing speakers enhances dialogue understanding and AI responses.

SD methodologies have traditionally been categorized into two approaches: modular and end-to-end systems. The modular approach involved distinct phases of segmentation (\citealp{Bredin23}; \citealp{Plaquet23}), feature extraction like x-vector \cite{8461375}, ECAPA-TDNN \cite{desplanques2020ecapa} or TitaNet \cite{koluguri2021titanet} , and clustering methods such as spectral clustering (SC) \cite{wang2022speaker} and agglomerative hierarchical clustering (AHC) \cite{scikit-learn}, each optimized for specific aspects of the SD process. On the other hand, end-to-end systems like MSDD \cite{park2022multiscale} or TOLD \cite{wang2023told} represented a more holistic approach, leveraging neural networks to perform diarization in an integrated manner, thereby streamlining the process and potentially enhancing accuracy. Despite all the advantages, audio-based approaches often suffered from low quality audio, similar acoustic features, and rapid change in speech causing faulty segmentation or clustering.

Recent advancements have explored the integration of semantic features to improve SD results \cite{park2023enhancing}. These attempts have ranged from utilizing text features directly \cite{Flemotomos_2020} to employing such models for error correction in post-processing stages (\citealp{paturi23_interspeech}; \citealp{wang2024diarizationlm}). Joint ASR+SD efforts \cite{kanda2022transcribetodiarize} and multimodal approaches \cite{cheng-etal-2023-exploring} have also been explored. However, these efforts in using semantic features have encountered several limitations: (a) The semantic features were often relied on outdated language models or were only used for a post-processing to rectify errors from preceding audio-based SD models, which did not fully exploit the semantic features. (b) There has been an absence of research exploring the use of text as the sole input for SD. This gap highlights a missed opportunity to fully explore the capabilities of semantic features in SD.

In response to these challenges, this paper proposes a novel text-based SD approach leveraging only the dialogue transcript as input. Our methodology is evaluated against a curated dataset and compared with multiple recent audio-based SD models. Our findings demonstrate that while our model excels in handling short conversations, it also delivers competitive performance in longer dialogues, challenging the conventional reliance on audio features.

\noindent The main contributions of this paper includes:
\begin{enumerate}
    \item Text-based sentence-level SD approaches using text as the only input that which achieves state-of-the-art result for short conversation.
    \item Data processing pipeline tailored to optimize SD on ASR-generated transcripts. 
    \item Comprehensive analysis about performance and error types of the text-based approach. 
\end{enumerate}

\section{Text-based Speaker Diarization}

We tackle the task of text-based SD as Sentence-level Speaker Change Detection in 2-speaker conversations. This approach prioritizes sentence-level analysis over word-level for its richer contextual information, which is more conducive to accurate speaker identification. Such granularity also addresses the label imbalance issue prevalent in word-level diarization, offering a more balanced dataset for model training and evaluation. 

2-speaker conversations represent a common and pragmatically significant scenario, making it an ideal focus for demonstrating the capabilities of text-based SD. Furthermore, the methodologies developed for dual-speaker contexts can be extended to multi-speaker scenarios with minimal adjustments as shown in Appendix \ref{sec:multispeaker-extension}.

\subsection{Single Prediction Model}
\label{sec:spm}
The single prediction model (SPM) operates by evaluating the probability of a speaker change between sentences, using surrounding utterances as context. Let $S=\{s_1, s_2, ..., s_n\}$ be a sequence of $n$ sentences in a conversation. The objective is to predict a binary variable $y_i$ for each pair of consecutive sentences $(s_i, s_{i+1})$, where $y_i=1$ if a speaker change occurs between $s_i$ and $s_{i+1}$, or $y_i=0$ otherwise. The change prediction result for $S$ will be a sequence of speaker change predictions $R=\{y_1, y_2, ..., y_{n-1}\}$ which can be used to deduce the final speaker information.

The model utilizes a context window of $h$ sentences at front and a context window of $k$ sentences at back of sentence $s_{i+1}$, where the context for the $i$-th prediction is defined as $C_i=\{s_{i-h+1}, s_{i-h+2}, ..., s_i, s_{i+1}, ..., s_{i+k+1}\}$. The prediction function $f$ can be expressed as $y_i=f(C_i, \theta)$, where $\theta$ represents the model parameters. $f$ can be implemented by any model that takes a sequence as an input and outputs a binary prediction label. The model is trained to minimize the loss function $L$, typically a binary cross-entropy loss, over all prediction points:

\begin{equation}
\label{eq:single-loss}
\begin{split}
    L(\theta)=-\frac{1}{n-1}\sum^{n-1}_{i=1}[y_i\log(f(C_i, \theta)) \\ +(1-y_i)\log(1-f(C_i, \theta))]
\end{split}
\end{equation}

\subsection{Multiple Prediction Model}
\label{sec:mpm}
To enhance accuracy and robustness, this paper introduces a multiple prediction model (MPM) that aggregates predictions over several windows within a dialogue so that more contextual information is leveraged on one prediction. The MPM extends the SPM by making predictions over multiple points within a sliding window across the conversation. Let $W=\{w_1, w_2, ..., w_m\}$ be a sequence of windows, where each window $w_j$ consists of a subsequence of sentences from $S$, and $m$ is the total number of windows covering the conversation. Each window $w_j$ overlaps with its predecessors and successors. The objective is to predict a sequence of binary variables $y_i$ for each window $w_j$, where each element of $y_i$ corresponds to a potential speaker change within $w_j$ between two consecutive sentences. The prediction for window $w_j$is given by: $y_j=g(w_j, \phi)$, where $g$ represents the multiple prediction function like $f$, and $\phi$ denotes the model parameters specific to the multiple prediction task. The model is trained to minimize a similar loss function $L'$, adjusted for multiple predictions:
\begin{equation}
\label{eq:multiple-loss}
\begin{split}
    L'(\phi)=-\frac{1}{\sum^{m}_{j=1}|w_j|-1}\sum^{m}_{j=1}\sum^{|w_j|-1}_{i=1} \\ [y_{ji}\log(g(w_j, \phi)_i)+(1-y_{ji})\log(1-g(w_j, \phi)_i)]
\end{split}
\end{equation}
where $y_{ji}$ is the $i$-th element of $y_j$, and $g(w_j, \phi)_i$ is the $i$-th prediction of $g$ for window $w_j$.

To consolidate predictions across overlapping windows, an aggregation mechanism is employed. For any given potential speaker change point $p$, predictions from all windows encompassing $p$ are aggregated to determine the final decision $Y_p$.

\begin{equation}
\label{eq:multiple-aggregate}
    Y_p=Aggregate(\{g(w_j, \phi)_p | p \in w_j\})
\end{equation}
The aggregation function could be a simple majority vote, a weighted average based on confidence scores, or another suitable method designed to optimize prediction accuracy.

\begin{figure}[htbp!]
\centering
\includegraphics[width=\columnwidth]{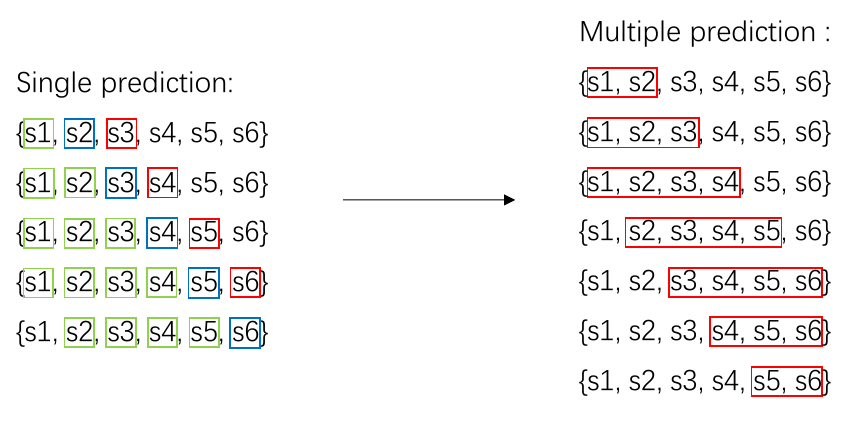}
\caption{Demonstration of SPM and MPM on a six-sentence conversation. 
}
\label{fig:spm-mpm-demo}
\end{figure}

\subsection{Data Processing}
Acknowledging the primary application of SD on ASR-generated transcripts, the training data for such text-based models should mimic this environment. The original audio recordings are processed through state-of-the-art ASR to produce transcripts, which were then aligned with ground-truth annotations to simulate real-world ASR discrepancies. This method ensures our model is fine-tuned for practical applications, particularly in improving the fidelity of ASR-generated transcripts for SD tasks.

\section{Experiment}
\subsection{Dataset}
This paper uses a curated dataset from 7 diverse, open-domain sources with both short and long conversations. The dataset is split with a ratio of 8:1:1 for train, development, and test set on conversation level to prevent data leakage. The details of the curated dataset are at Table \ref{tab:appendix-dataset} in Appendix \ref{sec:appendix-dataset}. 

\begin{figure*}[htbp!]
\centering
\includegraphics[width=\textwidth]{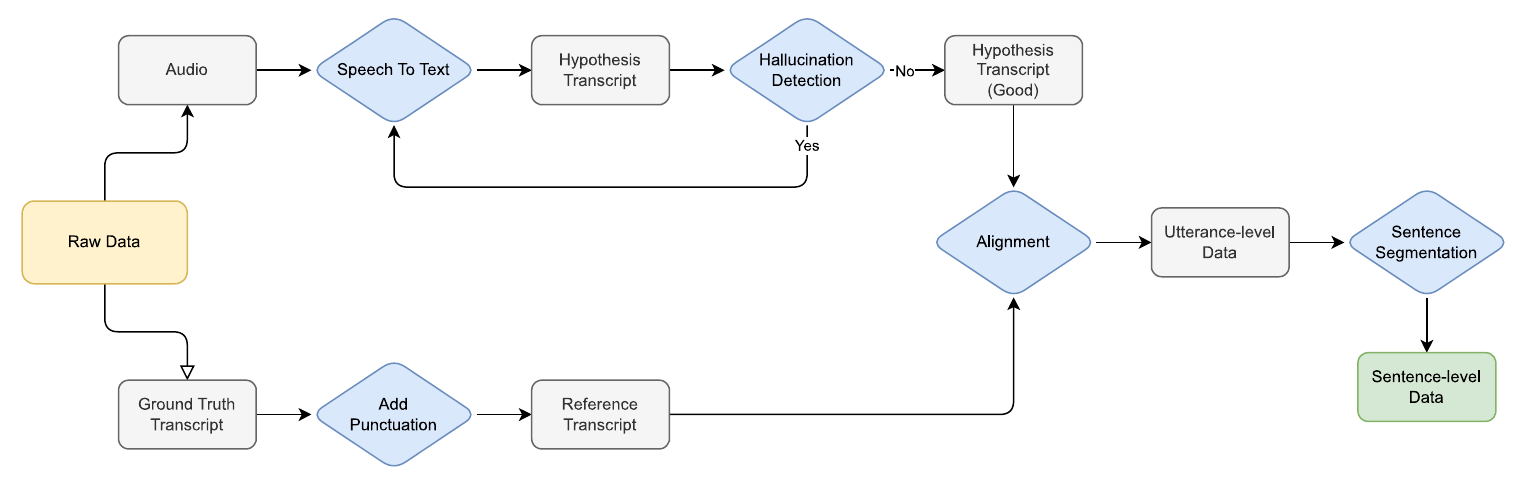}
\caption{Data processing procedure that uses reference transcript with ground truth speaker information and ASR-generated hypothesis transcript to obtain real training and evaluation data with ASR discrepancies.}
\label{fig:data-processing}
\end{figure*}

For data processing, OpenAI Whisper \cite{radford2022robust} is employed for transcription due to its robust performance, with temperature and beam search tuning applied to mitigate hallucination. Punctuation correction and insertion are handled by GPT-4, while sentence segmentation necessary for further processing is achieved using spaCy \cite{Honnibal_spaCy_Industrial-strength_Natural_2020}. The align4d \cite{10356536} tool is utilized for aligning ASR-generated transcripts with ground truth to map speaker information. This combination of Whisper, GPT-4, spaCy, and align4d forms a streamlined approach for preparing and refining speech transcription data for linguistic research, as depicted in Figure \ref{fig:data-processing}.

\subsection{Models}
For both SPM and MPM, the T5-3B \cite{10.5555/3455716.3455856} model is used for the experiment because of its flexibility in terms of input and output format, and its balance between performance and efficiency. 
The regular T5 is chosen over further fine-tuned versions such as FLAN-T5 because our approach does not closely align with the tasks these versions were trained on. Therefore, prompt tuning is not employed in our methodology. For aggregation mechanism, the majority voting is used.

The Word Diarization Error Rate (WDER) \cite{shafey2019joint} is used as the evaluation metric. To account for varying conversation lengths in the curated dataset, WDER-S is introduced as the weighted average of WDER, based on the number of sentences in each conversation, to more accurately reflect model performance in longer conversations. For audio models, the metrics are calculated by aligning speaker segments to reference transcript using word-level time stamps.

\subsection{Result}
\begin{table}[htbp!]
\centering\resizebox{\columnwidth}{!}{
\begin{tabular}{@{}ccccccc@{}} 
\toprule
\multirow{2}{*}{Model} & \multicolumn{2}{c}{$\leq$ 15 Min.} & \multicolumn{2}{c}{$>$ 15 Min.} & \multicolumn{2}{c}{Overall} \\ 
\cmidrule(lr){2-3} \cmidrule(lr){4-5} \cmidrule(lr){6-7}
& WD & WD-S & WD & WD-S & WD & WD-S \\
\midrule
pyannote & 26.9 & 23.3 & 13.7 & 12.7 & 22.5 & 18.7 \\
x-vector+SC & 37.8 & 33.9 & 15.0 & 17.5 & 30.2 & 18.4 \\
x-vector+AHC & 29.8 & 26.9 & 24.1 & 26.8 & 27.9 & 25.8 \\
ECAPA+SC & 40.2 & 37.1 & 19.9 & 15.2 & 33.4 & 27.8 \\
ECAPA+AHC & 29.1 & 25.6 & 16.6 & 26.7 & 24.9 & 23.9 \\
NeMo-TitaNet & 23.3 & 17.7 & 10.3 & 8.8 & 18.9 & 12.7 \\
NeMo-MSDD & 23.0 & 17.5 & 8.5 & 7.8 & 18.1 & 12.3 \\
TOLD & 20.6 & 12.9 & \bf 8.0 & \bf 6.9 & 16.4 & \bf 9.9 \\
\midrule
T5-3B SPM & 31.2 & 33.4 & 52.8 & 56.3 & 38.4 & 44.0 \\
T5-3B MPM & \bf 4.9 & \bf 5.5 & 11.4 & 12.9 & \bf 10.1 & 10.4 \\
\bottomrule
\end{tabular}}
\caption{Performance comparison in terms of WDER and WDER-S (\%) with audio-based SD systems (the lower the better) with respect to time. A more detailed conversation length analysis between the best model from audio and text are in Appendix \ref{sec:conversation-length}.}
\label{tab:sd-compare}
\end{table}

Our text-based models are compared to recent audio-based SD systems, including both modularized and end-to-end systems. The results in Table \ref{tab:sd-compare} indicate that text-based SD, especially with multiple predictions, offers a promising alternative to traditional audio-based methods, excelling in short conversational contexts.


\subsection{Length-based Analysis}
Both SPM and MPM are trained and evaluated with respect to different lengths of input context in terms of the total number of sentences. The experiment is conducted with lengths of {4, 6, 8} sentences, as shown in Table \ref{tab:sentence-compare}, due to the input length limitation of the T5 model and to ensure the aggregation mechanism is meaningful.

\noindent Table~\ref{tab:sentence-compare} shows that the performance increases as the length of input context increases, especially for longer dialogues as the WDER-S metric decreases drastically. This demonstrates the Large Language Models' (LLMs) ability to deduce speaker information from a long surrounding context.

\begin{table}[htbp!]
\centering\resizebox{\columnwidth}{!}{
\begin{tabular}{@{}ccccccc@{}}
\toprule
\multirow{2}{*}{Input Sentence} & \multicolumn{2}{c}{T5-3B SPM} & \multicolumn{2}{c}{T5-3B MPM} \\ 
\cmidrule(lr){2-3} \cmidrule(lr){4-5}
& WDER & WDER-S & WDER & WDER-S \\
\midrule
4 & 42.8 & 47.5 & 7.3 & 27.7 \\
6 & 38.8 & 42.9 & \bf 5.6 & 16.5 \\
8 & 38.4 & 44.0 & 10.1 & \bf 10.4 \\
\bottomrule
\end{tabular}}
\caption{Performance comparison of SPM and MPM with respect to number of input sentences.}
\label{tab:sentence-compare}
\end{table}

\subsection{Text-based Errors}
Three major types of inputs causing incorrect predictions are identified manually through random selection of 50 incorrect predictions: (a) short sentences, (b) similar speaker roles, and (c) grammatical errors. The examples and details of these types are in Appendix \ref{sec:text-based-error-example}. It is worth noting that these types of errors sometimes occur together, particularly in open-domain daily communications, which increases the difficulty for text-based SD. By identifying these types of errors, targeted strategies such as linguistic pattern recognition and error correction mechanisms can be developed for model improvement.

\subsection{Aggregation Efficacy}
To further explore the effectiveness of MPM over SPM through aggregation, the percentage of predictions that were originally wrong but revised through aggregation is calculated among all predictions that are incorrect at the same position at least once, as shown in Table \ref{tab:aggregation}. This result shows the effectiveness of aggregation mechanism in improving robustness. 

\begin{table}[htbp!]
\centering\small{ 
\begin{tabular}{c|c} 
\toprule
\bf Types of Prediction & \bf \% \\
\midrule
Partially Incorrect, Aggregated to Correct & 40.9 \\
Partially Incorrect, Aggregated to Incorrect & 18.0 \\
Consistently Incorrect & 41.1 \\
\bottomrule
\end{tabular}}
\caption{Percentage of three types of predictions with at least once of incorrect label in a window.}
\label{tab:aggregation}
\end{table}

\section{Conclusion}
This paper presents a novel text-based approach to Speaker Diarization (SD), focusing on the application of Sentence-level Speaker Change Detection (SCD) for identifying "who speaks what" in dialogues. By leveraging linguistic and semantic features in the text, the Single Prediction Model (SPM) and the Multiple Prediction Model (MPM) are proposed, which outperform traditional audio-based SD systems in short conversations. Our experiments on a curated dataset demonstrate the efficacy of our methodology, particularly the MPM. This research not only challenges the conventional reliance on acoustic features for SD but also shows the potential of text-based approaches in enhancing SD systems. As we look forward to expanding this work to multi-party dialogues, our findings highlight the possibilities of integrating semantic understanding into conversational AI, paving the way for more nuanced and contextually-aware diarization technologies.

\section{Limitations}
This study, while pioneering in its approach to text-based speaker diarization, encounters several limitations that are worth consideration for future research:
\begin{enumerate}
    \item The effectiveness of our text-based models heavily relies on the accuracy of ASR-generated transcripts. Mis-recognitions, omissions, or errors introduced during the ASR process can significantly affect diarization performance, particularly in noisy or challenging acoustic environments.
    \item Although our models show promising results in short conversations, their performance degrades as conversation length and complexity increase. The models struggle with maintaining context over long dialogues, suggesting a need for advanced mechanisms to handle extended interactions.
    \item The current study is conducted on English-language datasets, which may not capture the linguistic and cultural nuances of other languages. This limitation restricts the generalizability of our findings across diverse linguistic settings, potentially impacting the model's performance in multi-lingual or cross-cultural contexts.
    \item While this study focuses on 2-speaker conversations, real-world applications often involve multi-party interactions. Extending our approach to accommodate multiple speakers introduces additional complexity, requiring sophisticated methods to accurately identify and differentiate between a larger number of participants.
\end{enumerate}

\bibliography{anthology,custom}
\bibliographystyle{acl_natbib}

\appendix

\section{Appendix}
\label{sec:appendix}

\subsection{Dataset Details}
\label{sec:appendix-dataset}
Table \ref{tab:appendix-dataset} details the composition of our curated dataset, indicating the volume (in hours) and the number of dialogues sourced from each corpus. The selection was made to ensure a wide range of conversational contexts and settings, from formal meetings (AMI Corpus, ICSI Corpus) to casual conversations (CallFriend, CallHome English), and challenging acoustic environments (CHiME-5), providing a comprehensive base for training and evaluating conversational AI systems.

\begin{table}[htbp!]
\centering\resizebox{\columnwidth}{!}{
\begin{tabular}{c|cc}
\toprule
\bf Corpus & \bf Hour & \bf \# of Dialogue \\
\midrule
AMI Corpus \cite{10.1007/11677482_3} & 100 & 171 \\
CallFriend \cite{ldc96s46} & 20 & 41 \\
CallHome English \cite{ldc97s42} & 20 & 176 \\
CHiME-5 \cite{barker18_interspeech} & 50 & 20 \\
DailyTalk \cite{10095751} & 20 & 2541 \\
ICSI Corpus \cite{1198793} & 72 & 75 \\
SBCSAE \cite{SBCSAE} & 23 & 60 \\
\bottomrule
\end{tabular}}
\caption{Corpora used for the curated dataset.}
\label{tab:appendix-dataset}
\end{table}

\subsection{Conversation Length}
\label{sec:conversation-length}
To further analyze and compare the ability of text-based model on long conversation, this paper also compares the best of our text-based model (T5-3B MPM) to the best audio-based model (TOLD) with respect to the length of the conversation in the test set.

\begin{figure}[htbp!]
\centering
\begin{tikzpicture}
\begin{axis}[
    xlabel={Conversation Length (Min)},
    ylabel={WDER (\%)},
    xlabel style={at={(axis description cs:0.5,-0.15)},anchor=north,font=\small,inner sep=1pt,outer sep=1pt},
    ylabel style={at={(axis description cs:-0.15,.5)},anchor=south,font=\small,inner sep=1pt,outer sep=1pt},
    width=\columnwidth, 
    height=4.75cm, 
    xmin=0, xmax=6,
    ymin=0, ymax=25,
    xtick={0,1,2,3,4,5,6},
    xticklabels={0-5,5-10,10-15,15-20,20-25,25-30,30+},
    ytick={0,5,10,15,20,25},
    tick label style={font=\footnotesize},
    legend pos=north east,
    legend style={font=\tiny}, 
    ymajorgrids=true,
    grid style=dashed,
]

\addplot[
    color=blue,
    mark=o,
    ]
    coordinates {
    (0,23.57068637229433)(1,12.778907243515786)(2,9.71178302840618)(3,8.026575637778441)(4,5.8583136588228495)(5,5.861279069767442)(6,5.181617328279941)
    };
    \addlegendentry{TOLD (Audio)}

\addplot[
    color=red,
    mark=x,
    ]
    coordinates {
    (0,1.635378721887868)(1,6.936262989028277)(2,2.8031071982530742)(3,4.446680010285742)(4,15.52895503537822)(5,15.651468097120272)(6,14.34949401521916)
    };
    \addlegendentry{T5-3B MPM (Text)}

\end{axis}
\end{tikzpicture}
\caption{Average WDER with respect to Conversation Length for Audio and Text-based Groups}
\label{fig:conversation-length}
\end{figure}
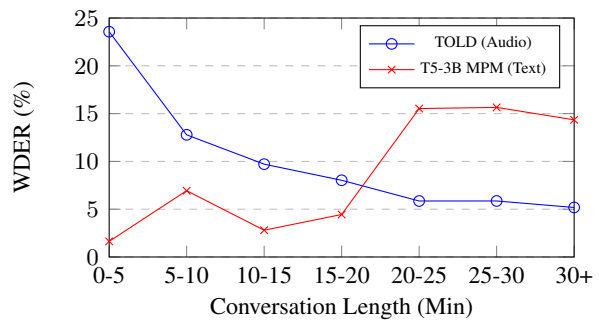

From Figure \ref{fig:conversation-length}, it is clear that the text-based model has a opposite trend in terms of SD performance with respect to length of conversation.

\subsection{Text-based Error Examples}
\label{sec:text-based-error-example}
\paragraph{Short Sentences} Short sentences often lack sufficient context, making it difficult for algorithms to accurately detect speaker changes. The brevity of responses minimizes the linguistic cues available for distinguishing between speakers.

\textbf{Example 1:}
\begin{itemize}
    \item \textbf{Dialogue}:
    \begin{quote}
    $s_1$: Wow. \\
    $s_2$: What time is it there? \\
    $s_3$: What time is it? \\
    $s_4$: It's 3:40. 
    \end{quote}
    
    \item \textbf{Model Prediction}: [A, B, B, A]
    \item \textbf{Correct Label}: [A, B, A, A]
\end{itemize}

The model fails to detect the change between $s_2$ and $s_3$ because all the sentences are short without much logical or linguistic information that can identify two speakers in this conversation.

\paragraph{Similar Speaker Roles} When participants have similar roles, their speech patterns often converge, making it challenging for models to discern speaker changes.

\textbf{Example 2:}
\begin{itemize}
    \item \textbf{Dialogue}:
    \begin{quote}
    $s_1$: They just said it was gonna be recorded whatever. \\
    $s_2$: So how's it going? \\
    $s_3$: Everything's going cool. \\
    $s_4$: When I first got here, things were kind of messed up, but I got your email.
    \end{quote}
    
    \item \textbf{Model Prediction}: [A, A, B, B]
    \item \textbf{Correct Label}: [A, B, A, A]
\end{itemize}

The model fails to detect the change between $s_1$, $s_2$ and $s_3$, $s_4$ because both of the speakers are teenage students who have experienced the same situation, who both have similar speaking patterns.

\paragraph{Grammatical Errors} Grammatical inaccuracies, such as typos and unconventional grammar, introduce ambiguity, complicating the detection of speaker changes.

\textbf{Example 3:}
\begin{itemize}
    \item \textbf{Dialogue}:
    \begin{quote}
    $s_1$: How things with you busy? \\
    $s_2$: I guess I sent you an email, but I suppose you haven't gotten it. 
    \end{quote}

    \item \textbf{Model Prediction}: [A, B]
    \item \textbf{Correct Label}: [A, A]
\end{itemize}

The model fails to detect the change between $s_1$ and $s_2$ because of the incorrect grammar in $s_1$.

\subsection{Multi-speaker Extension}
\label{sec:multispeaker-extension}
The MPM in Section \ref{sec:mpm} can be generalized to multi speaker situations. Let $L=\{l_1, l_2, ..., l_p\}$ be a set of possible speaker labels. For each window $w_j$, the model predicts a sequence of speaker labels $l_j$  for the sentences within $w_j$. The prediction for window $w_j$ is given by $l_j=h(w_j, \phi)$, where $h$ represents the prediction function for the multiple prediction model, and $\phi$ denotes the model parameters. The loss function $L''$, considering multiple speakers and windows, is expressed as:
\begin{equation}
\small
\label{eq:multispeaker-loss}
    L''(\phi)=-\frac{1}{\sum^{m}_{j=1}|w_j|}\sum^{m}_{j=1}\sum^{|w_j|}_{i=1}\sum^{p}_{k=1}y_{jik}\log(g(w_j, \phi)_{ik})
\end{equation}
where $y_{jik}$ is is a binary indicator if speaker $k$ is the correct speaker for the $i$-th sentence of $w_j$, and $g(w_j, \phi)_i$ is the is the predicted probability that the $i-th$ sentence in window $w_j$ is spoken by speaker $k$.

For multiple speakers, speaker label matching need to be done before actual aggregation. Such label matching can be done through bipartite matching between every consecutive two windows $w_j$. Through label matching, speakers labels across different windows can be unified for the whole conversation and normal aggregation mechanism can then be applied.

\end{document}